\documentclass[letterpaper, 10 pt, conference]{ieeeconf}  

\IEEEoverridecommandlockouts                              

\usepackage[english]{babel}
\usepackage{times} 
\usepackage{cite}
\usepackage[font=footnotesize]{subcaption}
\usepackage{calrsfs}
\usepackage[nolist]{acronym}

\usepackage{cellspace}
\newcolumntype{D}{>{\hfill}N{3}{2}<{\hfill}}

\usepackage{xcolor}
\usepackage[%
colorlinks = false,
linkcolor = blue,
urlcolor  = blue,
citecolor = blue,
anchorcolor = blue]{hyperref}

\usepackage{url}
\makeatletter
\g@addto@macro{\UrlBreaks}{\UrlOrds}
\makeatother

\usepackage{graphicx} 
\usepackage{epsfig} 
\usepackage{pgfplots}
\usepackage{caption}
\usepackage{subcaption}

\pgfplotsset{compat=1.15}
\graphicspath{{./figure/}}

\usepackage[export]{adjustbox}

\makeatletter
\let\MYcaption\@makecaption
\makeatother

\makeatletter
\let\@makecaption\MYcaption
\makeatother

\usepackage{mathptmx} 
\usepackage{amsmath} 
\usepackage{amssymb}  
\usepackage{nicefrac}
\usepackage{mathtools}
\usepackage{optidef}
\usepackage{leftidx}
\usepackage{bm} 
\DeclareMathAlphabet{\pazocal}{OMS}{zplm}{m}{n}

\DeclareMathOperator{\arctantwo}{arctan2}

\def\x#1{\texttt{\expandafter\string\csname#1\endcsname}&\expandafter$\csname#1\endcsname$}

\makeatletter%
\AfterPreamble{%
	\usepackage{hyperref}%
	\let\oldhypertarget\hypertarget%
	\renewcommand{\hypertarget}[2]{%
		\oldhypertarget{#1}{#2}%
		\protected@write\@mainaux{}{%
			\string\expandafter\string\gdef%
			\string\csname\string\detokenize{#1}\string\endcsname{#2}%
		}%
	}%
	\newcommand{\myhyperlink}[1]{%
		\hyperlink{#1}{\csname #1\endcsname}%
	}%
}
\makeatother%

\newcounter{Definition}
\newcounter{Theorem}
\makeatletter
\providecommand{\bigsqcap}{%
	\mathop{%
		\mathpalette\@updown\bigsqcup
	}%
}
\newcommand*{\@updown}[2]{%
	\rotatebox[origin=c]{180}{$\m@th#1#2$}%
}
\makeatother

\usepackage{tikz}
\pgfdeclarelayer{foreground}
\pgfsetlayers{background,main,foreground}
\usetikzlibrary{calc, trees, positioning, arrows, chains, shapes.geometric, quotes, angles, 
	decorations.pathreplacing, decorations.pathmorphing, shapes, decorations.markings, matrix, 
	shapes.symbols, automata, backgrounds, calc, patterns, shadows.blur, spy, fit}

\tikzstyle{vecArrow} = [thick, decoration={markings,mark=at position
	1 with {\arrow[semithick]{open triangle 60}}},
double distance=1.4pt, shorten >= 5.5pt,
preaction = {decorate},
postaction = {draw,line width=1.4pt, white,shorten >= 4.5pt}]

\usepackage{siunitx}
\DeclareSIUnit\pixel{pixel}

\usepackage{todonotes}

\usepackage{algorithm}
\usepackage[noend]{algpseudocode}

\makeatletter
\def\BState{\State\hskip-\ALG@thistlm}
\makeatother

\title{\LARGE \bf
	A Vision-Based Algorithm for a Path Following Problem 
}

\author{Mario Terlizzi$^{1}$, Giuseppe Silano$^{2}$, Luigi Russo$^{1}$, Muhammad Aatif$^{1}$,\\ 
Amin Basiri$^{1}$, Valerio Mariani$^{1}$, Luigi Iannelli$^{1}$, and Luigi Glielmo$^{1}$
	\thanks{This project was partially funded by the ECSEL Joint Undertaking (JU) research and 
	innovation programme AFarCloud and COMP4DRONES under grant agreement no. 783221 and no. 826610, 
	respectively, and by the European Union's Horizon 2020 research and innovation programme 
	AERIAL-CORE under grant agreement no. 871479. The JU receives support from the European Union's 
	Horizon 2020 research and innovation programme and Spain, Germany, Austria, Portugal, Sweden, 
	Finland, Czech Republic, Poland, Italy, Latvia, Greece, and Norway.}
	\thanks{$^{1}$Mario Terlizzi, Luigi Russo, Muhammad Aatif, Amin Basiri, Valerio Mariani, Luigi 
	Iannelli, and Luigi Glielmo are with the Department of Engineering, University of Sannio in 
	Benevento, Benevento, Italy (email:  {\tt\small \{mterlizzi, luirusso, maatif, basiri, 
	vmariani, luiannel, glielmo\}@unisannio.it}).}%
	\thanks{$^{2}$Giuseppe Silano is with the Faculty of Electrical Engineering, Czech Technical 
	University in Prague, Czech Republic (email: {\tt\small giuseppe.silano@fel.cvut.cz})}%
}

\begin{document}
	
\maketitle
\thispagestyle{empty}
\pagestyle{empty}


\begin{acronym}
	\acro{CoM}[CoM]{Center of Mass}
	\acro{FOV}[FoV]{Field of View}
	\acro{IBVS}[IBVS]{Image-Based Visual Servoing}
	\acro{IPS}[IPS]{Image Processing System}
	\acro{MAV}[MAV]{Micro Aerial Vehicle}
	\acro{NGL}[NGL]{Nonlinear Guidance Law}
	\acro{TP}[TP]{Trajectory Planning}
	\acro{PI}{Proportial-Integral}
	\acro{PP}[PP]{Path Planner}
	\acro{UAV}[UAV]{Unmanned Aerial Vehicle}
	\acro{VR}[VR]{Virtual Reality}
	\acro{VTP}[VTP]{Virtual Target Point}
	\acro{IMG}[IMG]{IMG}
\end{acronym}



\begin{abstract}
	
	A novel prize-winner algorithm designed for a path following problem within the~\ac{UAV} field 
	is presented in this paper. The proposed approach exploits the advantages offered by the pure 
	pursuing algorithm to set up an intuitive and simple control framework. A path for a 
	quad-rotor~\ac{UAV} is obtained by using downward facing camera images implementing 
	an~\ac{IBVS} approach. Numerical simulations in MATLAB\textsuperscript{\textregistered} 
	together with the MathWorks\textsuperscript{\texttrademark}~\ac{VR} toolbox demonstrate the 
	validity and the effectiveness of the proposed solution. The code is released as open-source 
	making it possible to go through any part of the system and to replicate the obtained results.
	
\end{abstract}







\section{Introduction}
\label{sec:introduction}

Path following is a relevant application problem within the~\acf{UAV} field. For instance, in 
precision agriculture scenarios having good path following algorithms is a fundamental requirement 
to preserve high productivity rates and plant growth~\cite{1_basso2019uav}. In civilian 
applications, monitoring of power lines can be encoded as a path following problem between several 
target regions that need to be inspected~\cite{Silano2021ICRARAL}. Whatever the field of 
application is, drones have to follow a path to accomplish the mission specifications safely and 
successfully.  

Looking at the path following problem, it can be divided into two parts: \textit{detection} and 
\textit{path following}~\cite{Dahroug2018MARSS, Rafique2020TAES}. As regards the \textit{path 
detection}, Hough transform~\cite{5_duda1972use} and its further 
developments~\cite{6_duan2010improved, Du2010TIMP} are considered the most valuable solutions in 
the literature to cope with the task. However, such a transformation demands a high computational 
effort making it hard to run on-board the aircraft when battery and processing constraints are 
tight. These requirements are increasingly stringent when considering~\acp{MAV}, where the sensor 
equipment and the vehicle dimensions are minimal. On the other hand, when the computational burden 
is at minimum, e.g., lightweight machine learning solutions have been recently proposed to tackle 
the problem~\cite{7_van2019ls, Tang2021PR}, the time required to set up the algorithms makes them 
difficult to apply in real world applications. For all such reasons, it is of interest to have low 
computational intensity algorithms, possibly without any prior knowledge of the surrounding 
environment, able to provide references to the \textit{path following} in a given time window.

Moving at the \textit{path following} level, much of the state-of-the-art 
solutions~\cite{8_sujit2013evaluation, 12_pelizer2017comparison} rely on the use of nonlinear 
guidance law~\cite{Keshmiri2018ICUAS}, vector field~\cite{Tuttle2021ARC}, and pure 
pursuit~\cite{Baqir2020IOP} algorithms due to their simple implementation and ease of use. Although 
the choice of path planner is application sensitive, general considerations can be provided. The 
performance of nonlinear guidance law degrades as the target acceleration changes rapidly 
introducing a not negligible delay in the trajectory generation. An adequate knowledge of the 
target velocity and acceleration is required to avoid instability issues~\cite{Keshmiri2018ICUAS}. 
On the other hand, a vector field solution prevents from such oscillations problems, but it is 
inherently characterized by a high computational effort~\cite{Tuttle2021ARC}. Besides, a pure 
pursuit approach is a suitable solution when tracking error and computational effort are critical. 
The position of a look-ahead point is set up at the beginning of the mission, and then updated at 
each step following some tracking criteria~\cite{Gautam2015ICUAS, 10_xavier2019path, 
Silano2019SMC}. The objective is to reduce the distance between the current position and the 
look-ahead position.

\begin{figure}
	\centering
	\includegraphics[width=0.35\textwidth]{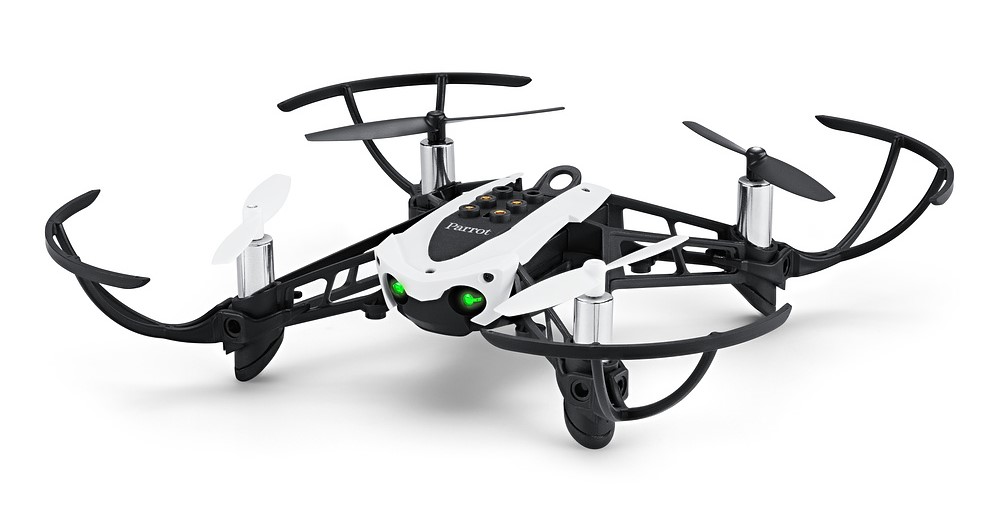}
	\caption{Parrot Mambo quad-rotor~\cite{15_Mathworks_url}.}
	\label{fig:parrotMambo}
\end{figure}

In this paper, we propose a novel winner-prize algorithm designed to deal with the path following 
problem in the context of the IFAC2020 MathWorks Minidrone competition~\cite{4_Mathworks_url}. The 
framework combines the advantages provided by the pure pursuit algorithm and some simple image 
processing to detect and to track a pre-established path. The lightweight and ease of 
implementation of the proposed solution allow the deployment on low computational 
capacity~\acp{MAV}, such as the Parrot Mambo~\cite{15_Mathworks_url} (see, 
Figure~\ref{fig:parrotMambo}) considered as a testbed for the application. Numerical simulations 
carried out in MATLAB together with the MathWorks~\acf{VR} toolbox~\cite{SilanoMATFly} show the 
validity and the effectiveness of the proposed approach. Moreover, the code is provided as 
open-source~\cite{GitHubCode} making it possible to go through any part of system and to replicate 
the obtained results. 

The paper is organized as follows. Section~\ref{sec:problemDescription} presents the problem, while 
in Sec.~\ref{sec:purPursuitTrackingAlgorithm} the vision-based path following algorithm is 
described. Numerical simulations are reported in Sec.~\ref{sec:simulationsResults}. Finally, 
conclusions are drawn in Sec.~\ref{sec:conclusions}. 



\section{Problem Description}
\label{sec:problemDescription}

\begin{figure}
	\centering
	\begin{tikzpicture}
	\node[anchor=south west,inner sep=0] (img) at (0,0) { 
		\includegraphics[width=0.35\textwidth]{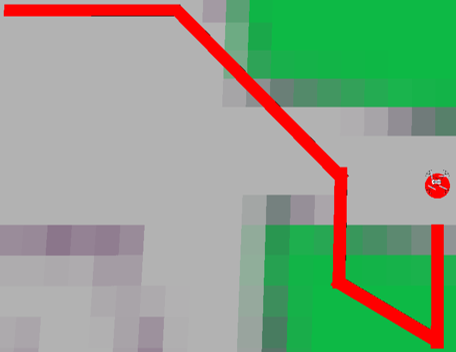}};
	\begin{scope}[x={(img.south east)},y={(img.north west)}]
	
	\draw [white, dashed, ultra thick] (0.95, 0.48) circle (0.065);
	
	\end{scope}
	\end{tikzpicture}
	\caption{Snapshot extracted from the virtual scenario. A dashed circle is used to indicated the 
	drone position.}
	\label{fig:arena}
\end{figure}

The work presented here finds an application within the IFAC2020 MathWorks Minidrone 
competition~\cite{4_Mathworks_url}, where the use of a model-based design approach is the aim of 
the specific contest. Path error and mission time are used as evaluation metrics for the algorithm. 
The whole process is the following: a quad-rotor~\ac{UAV} follows a pre-established red path by 
using its downward facing camera to get feedback from the environment. Images are updated according 
to the position and orientation of the vehicle simulated in the MATLAB~\ac{VR} world. No prior 
knowledge on the path and the surrounding scenario is given. The drone takes off and starts its 
motion looking at the path, and the mission stops with the recognition of an end-marker. At that 
time, the drone lands staying within the delimited area. Figure~\ref{fig:arena} shows the 
considered scenario. 





\section{Vision-Based Path Following Algorithm}
\label{sec:purPursuitTrackingAlgorithm}

\begin{figure}
	\centering
	\includegraphics[scale=0.5]{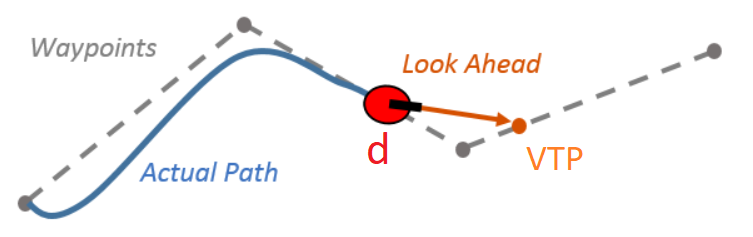}
	\caption{An illustrative example of the proposed vision-based path following algorithm works. 
	The red point $\mathbf{d}$ represents the drone position, while the orange point $\mathbf{w}$ 
	depicts the~\acs{VTP}.}
	\label{fig:pure pursuit}
\end{figure}

The vision-based path following algorithm combines the advantages offered by the pure pursuit 
algorithm~\cite{14_coulter1992implementation} with that of an easy image processing system to cope 
with the task. The algorithm starts selecting a target position ahead of the vehicle and that has 
to be reached, typically on a path. The framework is based on the following operations: (i) given 
the actual position $\mathbf{d}=(x_d, y_d)^\top \in \mathbb{R}^2$ where the~\ac{UAV} is located, 
a~\ac{VTP} is set over the track at $\mathbf{w}=(x_w, y_w)^\top \in \mathbb{R}^2$; then, (ii) the 
quad-rotor is commanded to reach the~\ac{VTP} along a straight line (essentially it is the pure 
pursuit algorithm with a curvature of infinite radius)~\cite{14_coulter1992implementation}, i.e., 
moving the vehicle from its current position to the goal position\footnote{The quad-rotor is 
assumed to fly at a fixed height along the entire mission.}. In Figure~\ref{fig:pure pursuit} an 
illustrative example of how the algorithm works is depicted. 

Contrary to the pure pursuit algorithm, the proposed approach exploits the intrinsic 
characteristics of the multi-rotor~\acp{UAV} motion: differently from ground vehicles with steering 
wheels, drones can follow a path without modifying their heading. Such an assumption allows 
reducing the time to accomplish the task by removing the control of the heading from the purposes 
of the path follower. 

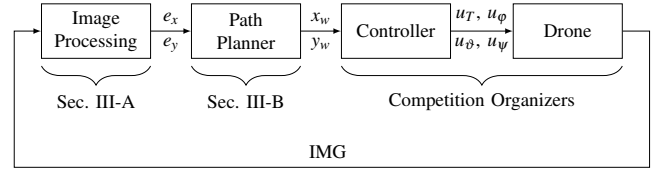
\begin{figure}
	\begin{center}
		\scalebox{0.725}{
			\begin{tikzpicture}
			
			\node (ImageProcessingSystem) at (-1.5,0) [draw, rectangle, minimum width=2cm, minimum 
			height=1cm, text centered, text width=5em]{Image Processing};
			
			\draw [decorate,decoration={brace, mirror, amplitude=10pt,raise=4pt},yshift=0pt] 
			(-2.5,-0.5) -- (-0.5,-0.5) node [black,midway,xshift=0.0cm,yshift=-0.8cm] {Sec.~III-A};
			
			\node (PathPlanner) at (1.25,0) [draw, rectangle, minimum width=2cm, minimum 
			height=1cm, text centered, text width=5em]{Path Planner};
			
			\draw [decorate,decoration={brace, mirror, amplitude=10pt,raise=4pt},yshift=0pt] 
			(0.25,-0.5) -- (2.25,-0.5) node [black,midway,xshift=0.0cm,yshift=-0.8cm] {Sec.~III-B};
			
			\node (Controller) at (4.00,0) [draw, rectangle, minimum width=2cm, minimum height=1cm, 
			text centered, text width=5em]{Controller};
			
			\node (Drone) at (7.15,0) [draw, rectangle, minimum width=2cm, minimum height=1cm, text 
			centered, text width=5em]{Drone};
			
			\draw [decorate,decoration={brace, mirror, amplitude=10pt,raise=4pt},yshift=0pt] 
			(3.00,-0.5) -- (8.15,-0.5) node [black,midway,xshift=0.0cm,yshift=-0.8cm] {Competition 
			Organizers};
			
			\draw[-latex] (ImageProcessingSystem) -- node[above]{$e_x$} node[below]{$e_y$} 
			(PathPlanner);
			\draw[-latex] (PathPlanner) -- node[above]{$x_w$} node[below]{$y_w$} (Controller);
			\draw[-latex] (Controller) -- node[above]{$u_T$, $u_\varphi$} 
			node[below]{$u_\vartheta$, $u_\psi$} (Drone);
			\draw[-latex] (Drone) -- ($ (Drone) + (1.5,0)$) -- ($ (Drone) + (1.5,-2.5)$) -- ($ 
			(ImageProcessingSystem) - (1.5,2.5)$) -- ($ (ImageProcessingSystem) - (1.5,0)$) --  
			(ImageProcessingSystem);
			
			\node at ($ (PathPlanner) - (-1.5,2.45)$) [text centered, above]{IMG};
			
			\end{tikzpicture}
		}
	\end{center}
	\caption{Control system architecture. From left to right: the image processing, path planner, 
	controller, and drone blocks.}
	\label{fig:block diagram}
\end{figure}

In Figure~\ref{fig:block diagram} the whole control scheme architecture is reported. The algorithm 
is mainly divided into two parts: (i) the~\ac{IPS} deals with extracting the red path from the 
camera images, providing the errors along the $x$- and $y$-axis of the camera frame between the 
current drone position and the~\ac{VTP} point, and recognizing the End-Marker for the landing 
operations; while, (ii) the~\ac{PP} figures out the path following problem by computing the new 
position $\mathbf{w}$ of the drone in the world frame\cite[Sec.~V]{SilanoMATFly} implementing 
an~\acf{IBVS} scheme. The control algorithm computes the commands $u_T$, $u_\varphi$, 
$u_\vartheta$, and $u_\psi$ that should be given to the drone in order to update its position and 
orientation in accordance to the~\ac{PP} references. 

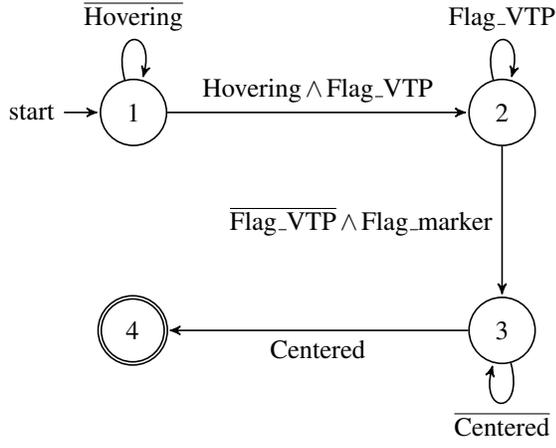
\begin{figure}
	\centering
	\begin{tikzpicture}[->, >=stealth', shorten >=0.8pt, auto, semithick, initial text=${\rm 
	start}$]
	
	\tikzstyle{every state}=[fill=white,text=black]
	
	\node[initial,state] (A)                    {$1$};
	\node[state]         (B)        [ right= 4cm of A] {$2$};
	\node[state]         (C) [ below= 2cm of B] {$3$};
	\node[accepting,state](D) [ left= 4cm of C] {$4$};
	
	\path (A) edge              node {${ \rm Hovering} \wedge {\rm Flag\_VTP}$} (B)
	edge [loop above] node {{${ \rm \overline{Hovering}}$}} (A)
	(B) edge       [left]       node {${ \rm \overline{Flag\_VTP} \wedge Flag\_marker}$} (C)
	edge [loop above] node {${ \rm Flag\_VTP}$} (B)
	(C) edge              node {${\rm Centered}$} (D)
	edge [loop below] node {${ \rm \overline{Centered}}$} (C);
	
	\end{tikzpicture}
	\caption{State machine implemented. State 1: \textit{Take-off}. State 2: \textit{Following}. 
	State 
		3: \textit{End-Marker}. State 4: \textit{Landing}.} 
	\label{fig:state machine}
\end{figure}

The overall mission is divided into four parts: \textit{Take off}, \textit{Following}, 
\textit{End-Marker}, and \textit{Landing}. A decision-making process has been implemented to 
achieve the competition objectives triggering the system from a state to another, as depicted in 
Figure~\ref{fig:state machine}. For each frame, the~\ac{IPS} accesses the system status and plan 
the next action (i.e., landing, following, etc.). The drone starts taking off from its initial 
position looking at the path. Once the vehicle reaches the hovering position, the~\ac{IPS} detects 
the path and the state machine enters in the \textit{Following} state, hence the path following 
starts. As soon as the~\ac{IPS} detects the End-Marker, the state machine exits from the 
\textit{Following} state and goes into the \textit{End-Marker} state. At this stage the mission 
stops, and the drone starts the landing. In the following subsections the implementation of the 
image processing system and path planner modules are detailed.



\subsection{Image processing system}
\label{sec:imageProcessingSystem}

Starting from the camera frames, the~\acl{IPS} takes care of separating the features of the 
pre-established path from that of the environment. 

\begin{figure}
	\centering
	\includegraphics[width=0.5\textwidth]{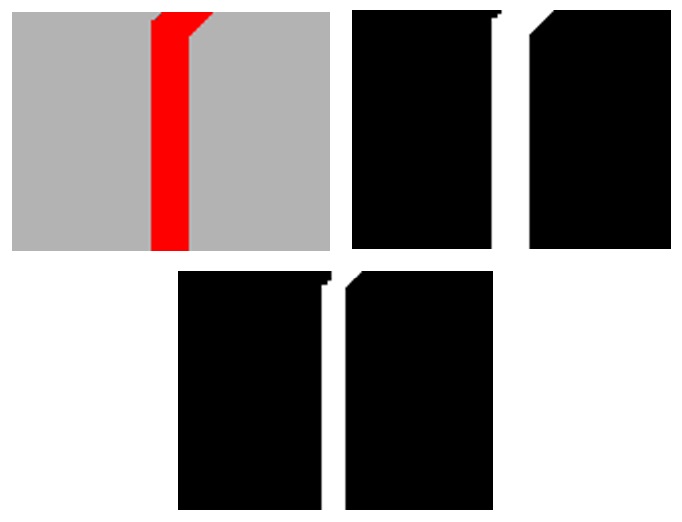}
	\caption{Original frame (upper left),  converted and binarized frame (upper right), and eroded 
	frame (lower).}
	\label{fig:preprocessing}
\end{figure}

The~\ac{IPS} receives frames of width $W$ and height $H$ from the camera sensor at each   
$T_\mathrm{IPS} = \SI{0.2}{\second}$, i.e., the camera sampling time. The image format is RGB with 
$8$ bits for each color channel. The path is \SI{0.01}{\meter} in width, while the landing marker 
is circular with a diameter of \SI{0.02}{\meter}. The path color is red, and this information is 
taken into consideration in all the elaborations to filter out the background scenario. The 
procedure consists of the following steps: first, the RGB frame is converted into an intensity 
level frame representation as follows 
\begin{equation}
	F(n,m)=f_R(n,m) -  \frac{f_G(n,m)}{GG} - \frac{f_B(n,m)}{GB} ,
\end{equation}
where the pair $(n,m)$ represents the pixel located at row $n \in \{1, 2, \dots, H\}$ and column $m 
\in \{1, 2, \dots, W\}$ of the image frame and $f_i$, with $i \in \{R, G, B\}$, provides the 
intensity level representation of the corresponding red, green and blue channels. An heuristic 
approach was used to tune the $GG$, $GB \geq 1$ parameter values. These parameters help to detect 
the pixels belonging to the path. Further, a binarization process based on a $K_T$  threshold value 
refines the process removing artifacts from the elaboration. The binarized frame can be described 
by the binary function $F_\mathrm{bin} \colon (n,m) \to \{0, 1\}$ whose output is one when the 
pixel belongs to the path and zero otherwise. Finally, an erosion operation is performed through a 
square kernel to shrink and regularize the binarized frame. In Figure~\ref{fig:preprocessing} the 
overall process is reported for a single sample frame.

Then, the obtained reference path is used in a twofold way: (i) to identify a new~\ac{VTP} 
belonging to the track; (ii) to detect the landing marker. The two tasks are described in the 
pseudocode reported in Algorithm~\ref{alg:imageProcessing}.

Looking at the algorithm, the first three functions (i.e., \texttt{channelConv}, 
\texttt{binarization}, and \texttt{erosion}) take care of extracting the path information from the 
frame. Then, the \texttt{detectTrack} and \texttt{detectMarker} functions deal with raising a flag 
when the path (\texttt{Flag\_VTP}) or the End-Marker (\texttt{Flag\_marker}) are detected. The path 
following algorithm starts with the~\ac{IPS} that computes the errors ($e_x$ and $e_y$) between the 
drone position and the~\ac{VTP} point for the~\ac{PP} by using a circular arc mask centered in the 
drone~\ac{CoM}\footnote{The assumption that the~\ac{CoM} being in the center of the reference 
frame, i.e., $x_\mathrm{CoM} = \nicefrac{H}{2}$ and $y_\mathrm{CoM}=\nicefrac{W}{2}$, is taken into 
consideration.} with thickness $R_\mathrm{max} - R_\mathrm{min}$\footnote{$R_\mathrm{max}$ and 
$R_\mathrm{min}$ are the outer and inner radius, respectively.}. 
\begin{algorithm}
	\caption{Image Processing System}\label{alg:imageProcessing}
	\begin{algorithmic}[1]
		\State $\text{IMG}  \gets \text{channelConv(\text{IMG})}$, \\
		$\text{IMG} \gets \text{binarization(\text{IMG})}$, \\
		$\text{IMG} \gets \text{erosion(\text{IMG})}$, \\
		$\text{Flag\_VTP} \gets \text{detectTrack(\text{IMG})}$, \\
		$\text{Flag\_marker}$ $\gets \text{detectMarker(\text{IMG})}$
		
		\If {$\text{Flag\_VTP}$} \\ 
		\quad $x_\mathrm{VTP}$, $y_\mathrm{VTP} \gets \text{vtp(frame)}$ \\
		\quad $e_x \gets x_\mathrm{VTP} - x_\mathrm{CoM}$ \\
		\quad $e_y \gets y_\mathrm{VTP} - y_\mathrm{CoM}$
		\Else \If{$\text{Flag\_marker}$} \\
		\quad \quad $\;$ $x_\mathrm{MARK}$, $y_\mathrm{MARK} \gets \text{cgMarker(frame)}$ \\
		\quad \quad $\;$ $e_x \gets x_\mathrm{MARK} - x_\mathrm{CoM}$ \\ 
		\quad \quad $\;$ $e_y \gets y_{\rm MARK} - y_{\rm CoM}$
		\EndIf \EndIf
		
		\State \Return $e_x$, $e_y$, $\text{Flag\_VTP}, \text{Flag\_marker}$ 
	\end{algorithmic}
\end{algorithm}


In Figure~\ref{fig:Arc_mask}, the arc mask considering the~\ac{VTP} position at time $\mathbf{t}_k$ 
is depicted, where $\mathbf{t}_k$ denotes the $k$-element of the time interval vector defined as 
$\mathbf{t} =(0, T_\mathrm{IPS}, \dots, NT_\mathrm{IPS})^\top \in \mathbb{R}^{N+1}$, with $k \in 
\mathbb{N}_0$. The orientation angle \mbox{$\vartheta = \arctantwo(x_\mathrm{VTP},y_\mathrm{VTP})$} 
is calculated with respect to the frame coordinates, where the $\arctantwo$ function is the 
four-quadrant inverse of the tangent function. A portion~$\varTheta$ of the arc mask is established 
by taking into account the previous~\ac{VTP}'s orientation. In particular, we set up two semi-arcs 
with width~$\nicefrac{\varTheta}{2}$, namely~\ac{FOV}, in counter-clockwise and clockwise 
directions from $\vartheta$. Then, the arc mask is applied to the eroded image obtaining 
the~\ac{VTP} point at $\mathbf{t}_{k+1}$. The function~\ac{VTP} calculates $x_\mathrm{VTP}$, 
$y_\mathrm{VTP}$, and $\vartheta$ which represent the frame coordinates and angle orientation of 
the~\ac{VTP} at  $\mathbf{t}_{k+1}$, respectively. Subsequently the corresponding errors with 
respect to the center of mass, i.e., $e_x$ and $e_y$, are computed inside the frame coordinates. 
Finally, the \texttt{Flag\_VTP} and the $e_x$ and $e_y$ values are provided as input to the~\ac{PP} 
at each $T_\mathrm{IPS}$. Figure~\ref{fig:track} shows the result of the entire process setup. 

\begin{figure}
	\begin{center}
		\scalebox{1.25}{
			\begin{tikzpicture}
			
			\draw[-latex] (-1.00,0) -- (2.5,0) node[below]{$x$};
			\draw[-latex] (0,-1.0) -- (0,2.5) node[left]{$y$};
			
			\draw[line width=2.15pt] (1.5,0.25) arc(10:80:1.5);
			
			\draw[-] (0,0) -- (1.05,1.05);
			
			\draw (0,0) arc(0:20:0.3);
			\draw[-latex] (0.84,0.15) arc (12:50:0.75cm);
			\node at (0.70,0.75) [below right]{\small $\nicefrac{\varTheta}{2}$}; 
			\draw[-latex] (0.10,0.44) arc (82.5:52:0.5cm);
			\node at (0.70,0.70) [left]{\small $\nicefrac{\varTheta}{2}$}; 
			\draw[-latex, cyan] (0.64,0) arc (0:55:0.5cm);
			\node at (0.40,0.05) [below right, cyan]{\small $\vartheta$}; 
			
			\draw[dashed] (0,0) -- (0.26,1.47); 
			\draw[dashed] (0,0) -- (1.47,0.26); 
			\draw[fill=red] (-0.08,0) arc(-180:180:0.1); 
			\node at (0,0) [below right]{\textcolor{red}{\small{d}}};
			\draw[fill=green] (0.975,1.075) arc(-180:180:0.1); 
			\draw (1.05, 1.05) node[above right]{\textcolor{green}{\small{VTP}}};
			
			\draw[dotted] (-1.30, -0.80) -- (1.25,1.55);
			\draw[dotted] (-0.40, -1.10) -- (2.15,1.35);
			\fill[blue!25, nearly transparent] (-1.30, -0.80) -- (1.25,1.55) -- (2.15,1.35) -- 
			(-0.40, -1.10) -- cycle;
			
			\end{tikzpicture}
		}
	\end{center}
	\caption{Arc mask. The drone position (red), the previous~\ac{VTP} (green), and the 
	pre-established path to follow (purple) are reported.} 
	\label{fig:Arc_mask}
\end{figure}
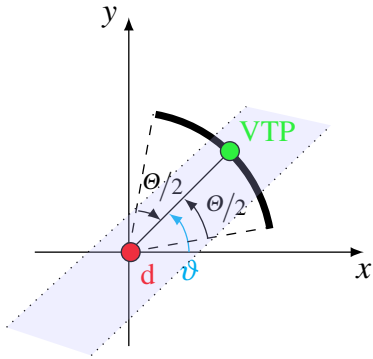
\begin{figure}
	\centering
	\includegraphics[width=0.5\textwidth]{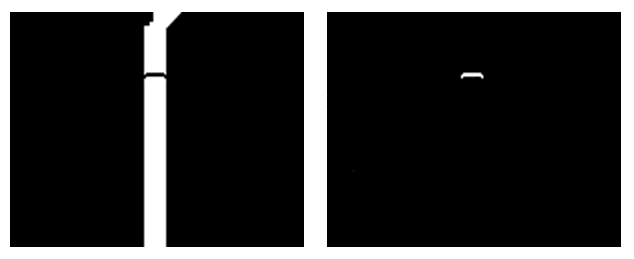}
	\caption{Frame after the application of the Arc mask (left). Extracted pixels belonging to the 
	path (right).}
	\label{fig:track}
\end{figure}

\begin{figure}
	\centering
	\includegraphics[width=0.5\textwidth]{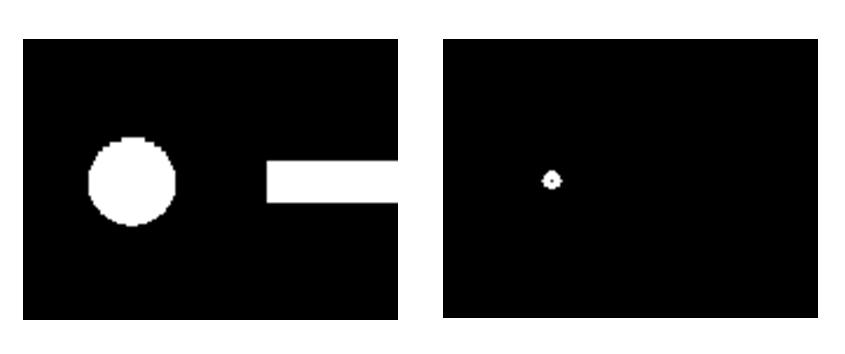}
	\caption{Original (left) and eroded frames (right) of the End-Marker are reported.}
	\label{fig:End_marker}
\end{figure}

It is worth noticing that when the landing marker is detected and no other~\ac{VTP} point is found 
in the frame, the~\ac{IPS} triggers the state machine in the \texttt{End-Marker} state. Here, the 
new main task of the~\ac{IPS} is to obtain the position of the End-Marker within the frame 
coordinates. An additional erosion process is performed by using a circular kernel, as depicted in 
Figure~\ref{fig:End_marker}. 

\subsection{Path planner}
\label{sec:pathPlanner}

The~\acl{PP} is designed to compute the position of the~\ac{VTP} point $\mathbf{w} = (x_w, 
y_w)^\top$ maintaining a constant altitude ($z_H$) while following the path. Roughly speaking, 
the~\ac{PP} computes the spatial coordinates $x_w$ and $y_w$ trying to reduce the errors, i.e., 
$e_x$ and $e_y$, between the drone position and the~\ac{VTP}. These values are later used by the 
drone controller to tune the command signals $u_T$, $u_\varphi$, $u_\vartheta$, and $u_\psi$, as 
described in Figure~\ref{fig:block diagram}. The proposed path planner is based on~\acl{PI} control 
loops. As a common rule in cascade structure, the inner loop, i.e., the~\ac{PP}, is regulated at a 
rate faster than the outer loop, i.e., the~\ac{IPS}. In our case, the~\ac{PP} runs at 
$\SI{200}{\hertz}$ ($T_\mathrm{PP} = \num{5e-3} \si{\second}$) while the~\ac{IPS} runs at 
$\SI{2}{\hertz}$ ($T_\mathrm{IPS} = \SI{0.2}{\second}$). These are a standard solution in the 
literature for quad-rotors control design~\cite{Dief2015}.

As described in Sec.~\ref{sec:purPursuitTrackingAlgorithm}, the path following stops with the 
detection of the End-Marker. At that time, the~\ac{IPS} implements a toggle switch behavior raising 
the \texttt{Flag\_marker} flag while holding low the \texttt{Flag\_VTP} flag. This mutually 
separates the \textit{Following} and \textit{Landing} phases avoiding instability issues. The 
pseudocode of the proposed algorithm is reported in Algorithm~\ref{alg:pathPlanner} with parameter 
values detailed in Table~\ref{tab:parameters}.

In Appendix~\ref{appendix}, we show how $\alpha$ can be set to control the velocity of vehicle 
along the entire mission. Therefore, the proposed Vision-Based Path Following algorithm makes it 
possible not only to generate the spatial coordinates $x_w$ and $y_w$ using a~\ac{IBVS} scheme but 
also to set the velocity during the entire mission.

\begin{algorithm}
	\caption{Path Planner}\label{alg:pathPlanner}
	\begin{algorithmic}[1]
		\State $e_x$, $e_y$, $\text{Flag\_VTP}$, $\text{Flag\_marker}$ 
		
		\If{$\text{Flag\_VTP}$} \\
		\quad $x_{k+1} \gets x_k + \alpha e_x$\\
		\quad $y_{k+1} \gets y_k + \alpha e_y$\\
		\quad $z_{k+1} \gets  z_H$
		\EndIf
		
		\If{$\text{Flag\_marker}$}
		\If{$(e_x=0 \wedge e_y=0)$} \\
		\quad \quad $\;$ $x_{k+1}  \gets x_k $\\
		\quad \quad $\;$ $y_{k+1} \gets y_k $\\
		\quad \quad $\;$ $z_{k+1} \gets 0$
		\Else \\
		\quad \quad $\;$ $x_{k+1} \gets x_k + \beta e_x$\\
		\quad \quad $\;$ $y_{k+1}  \gets y_k + \beta e_y$\\
		\quad \quad $\;$ $z_{k+1} \gets z_H$
		\EndIf \EndIf
		
		\State $x_w \gets x_{k+1}$, $y_w \gets y_{k+1}$, $z_w \gets z_{k+1}$
		
		\State \Return $x_w$, $y_w$, $z_w$
	\end{algorithmic}
\end{algorithm} 




\section{Numerical Results}
\label{sec:simulationsResults}

To demonstrate the validity and effectiveness of the proposed framework, numerical simulations have 
been carried out by using the 2019b release of MATLAB equipped with MathWorks Virtual Reality 
toolbox~\cite{16_Mathworks_url} and Parrot support package for Simulink~\cite{15_Mathworks_url}. 
The video available at~\cite{YouTubeVideo} illustrates in a direct way how the system works, i.e., 
the ability of the quad-rotor~\ac{UAV} to follow the pre-established red path and to land on the 
End-Marker. In addition, the video shows the behavior of the~\ac{IPS} and~\ac{PP} that never lose 
the path during the entire mission. 

\begin{figure}
	\begin{center}
		\hspace{-0.725cm}
		\begin{subfigure}[c]{0.45\columnwidth}
			\scalebox{0.52}{
				\begin{tikzpicture}
				\begin{axis}[%
				width=2.8119in,%
				height=1.8183in,%
				at={(0.758in,0.481in)},%
				scale only axis,%
				xmin=-3.7,%
				xmax=0,%
				ymax=3,%
				ymin=-0.25,%
				xmajorgrids,%
				ymajorgrids,%
				ylabel style={yshift=0cm}, 
				xlabel={X [\si{\meter}]},%
				ylabel={Y [\si{\meter}]},%
				axis background/.style={fill=white},%
				legend style={at={(0.725,0.875)},anchor=north,legend cell align=left, draw=none, 
					draw=white!15!black}
				]
				\addplot [color=blue, solid, line width=1.15pt] 
				file{matlabPlots/track_downsampled.txt};%
				\addplot [color=red, dashed, line width=1.15pt] 
				file{matlabPlots/alpha_005_downsampled.txt};%
				\legend{$\text{path}$, $\alpha = 0.05\text{,}\, T_s = \SI{30}{\second}$};%
				\end{axis}
				\end{tikzpicture}
			}
			\caption{}
		\end{subfigure}
		\hspace{0.25cm}
		\begin{subfigure}[c]{0.45\columnwidth}
			\scalebox{0.52}{
				\begin{tikzpicture}
				\begin{axis}[%
				width=2.8119in,%
				height=1.8183in,%
				at={(0.758in,0.481in)},%
				scale only axis,%
				xmin=-3.7,%
				xmax=0,%
				ymax=3,%
				ymin=-0.25,%
				xmajorgrids,%
				ymajorgrids,%
				ylabel style={yshift=0cm}, 
				xlabel={X [\si{\meter}]},%
				ylabel={Y [\si{\meter}]},%
				axis background/.style={fill=white},%
				legend style={at={(0.725,0.875)},anchor=north, legend cell align=left, draw=none, 
					draw=white!15!black}
				]
				\addplot [color=blue, solid, line width=1.15pt] 
				file{matlabPlots/track_downsampled.txt};%
				\addplot [color=green, dashed, line width=1.15pt] 
				file{matlabPlots/alpha_004_downsampled.txt};%
				\legend{$\text{path}$, $\alpha = 0.04\text{,}\, T_s = \SI{34}{\second}$};%
				\end{axis}
				\end{tikzpicture}
			}
			\caption{}
		\end{subfigure}
		\\
		\vspace{0.05cm}
		\hspace{-0.75cm}
		\begin{subfigure}[c]{0.45\columnwidth}
			\scalebox{0.52}{
				\begin{tikzpicture}
				\begin{axis}[%
				width=2.8119in,%
				height=1.8183in,%
				at={(0.758in,0.481in)},%
				scale only axis,%
				xmin=-3.7,%
				xmax=0,%
				ymax=3,%
				ymin=-0.25,%
				xmajorgrids,%
				ymajorgrids,%
				ylabel style={yshift=0cm}, 
				xlabel={X [\si{\meter}]},%
				ylabel={Y [\si{\meter}]},%
				axis background/.style={fill=white},%
				legend style={at={(0.725,0.875)},anchor=north,legend cell align=left, draw=none, 
				draw=white!15!black}
				]
				\addplot [color=blue, solid, line width=1.15pt] 
				file{matlabPlots/track_downsampled.txt};%
				\addplot [color=yellow, dashed, line width=1.15pt] 
				file{matlabPlots/alpha_003_downsampled.txt};%
				\legend{$\text{path}$, $\alpha = 0.03\text{,}\, T_s = \SI{47}{\second}$};%
				\end{axis}
				\end{tikzpicture}
			}
			\caption{}
		\end{subfigure}
		\hspace{0.25cm}
		\begin{subfigure}[c]{0.45\columnwidth}
			\scalebox{0.52}{
				\begin{tikzpicture}
				\begin{axis}[%
				width=2.8119in,%
				height=1.8183in,%
				at={(0.758in,0.481in)},%
				scale only axis,%
				xmin=-3.7,%
				xmax=-3.4,%
				ymax=3,%
				ymin=2,%
				xmajorgrids,%
				ymajorgrids,%
				ylabel style={yshift=0cm}, 
				xlabel={X [\si{\meter}]},%
				ylabel={Y [\si{\meter}]},%
				axis background/.style={fill=white},%
				legend style={at={(0.475,0.945)},anchor=north,legend cell align=left, draw=none, 
				legend columns=-1, draw=white!15!black}
				]
				\addplot [color=red, dashed, line width=1.15pt] 
				file{matlabPlots/alpha_005_downsampled.txt};%
				\addplot [color=green, dashed, line width=1.15pt] 
				file{matlabPlots/alpha_004_downsampled.txt};%
				\addplot [color=yellow, dashed, line width=1.15pt] 
				file{matlabPlots/alpha_003_downsampled.txt};%
				\legend{$\alpha = 0.05$, $\alpha = 0.04$, $\alpha = 0.03$};%
				\addplot [color=blue, solid, line width=1.15pt] 
				file{matlabPlots/track_downsampled.txt};%
				\end{axis}
				\end{tikzpicture}
			}
			\caption{}
			\label{subfig:comparisonPlot}
		\end{subfigure}
	\end{center}
	\caption{Trajectory plots. From left to right: the desired and the drone paths for various 
	values of $\alpha$ are represented. The mission time $T_s$ and a comparison between the 
	considered $\alpha$ values are also reported.}
	\label{fig:plot_trajectory}
\end{figure}
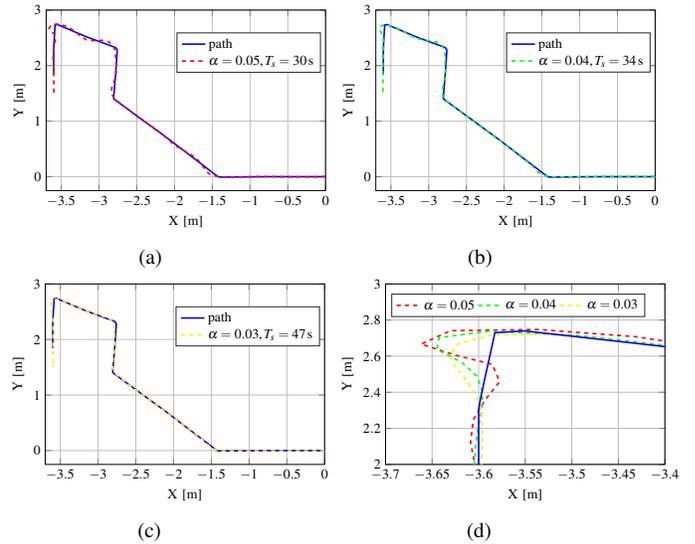

In Figure~\ref{fig:plot_trajectory} a comparison of the system performance by using various values 
of $\alpha$ is reported. As can be seen from the plots, the larger $\alpha$ is, the lower the 
mission time ($T_s$) is. On the other hand, the lower the mission time is, the greater the path 
error is. Looking at the zoom plot (see, Figure~\ref{subfig:comparisonPlot}) it is even clearer how 
the system performance degrades with increasing $\alpha$ value, and these are all the more evident 
as the path is angular. For the considered scenario, an heuristic approach was used to tune the 
$\alpha$ and $\beta$ parameter values.


\begin{table}
	\centering
	\caption{Parameter values.}
	\label{tab:parameters}
	\begin{tabular}[t]{|l|l|c|c|}
		\hline
		
		& \textbf{Sym} & \textbf{Value}\\
		
		\hline
		
		Sampling time & $T_{\rm PP}$ & \num{5e-3} \si{\second}\\
		Sampling time & $T_{\rm IPS}$ & 0.2 \si{\second}\\
		PP constant & $\beta$ & \num{18e-3} \si{\meter \per \pixel}\\
		Frame height & $H$ & 120 \si{\pixel}\\
		Frame width & $W$ & 160 \si{\pixel}\\
		Outer radius Arc mask & $R_{\rm max}$ & 28 \si{\pixel} \\
		Inner radius Arc mask & $R_{\rm min}$ & 26 \si{\pixel} \\
		IPS constant & $GB$ & 2 \\
		IPS constant & $GG$ & 2 \\
		IPS threshold & $K_T$ & 150 \\
		FOV Arc mask & $\vartheta$ & 2.3 \si{\radian} \\
		Drone height & $z_H$ & 1 \si{\metre} \\
		
		\hline
	\end{tabular}
\end{table}%


\begin{figure}
	\begin{center}
		\scalebox{0.9}{
			\begin{tikzpicture}
			\begin{axis}[%
			width=2.8119in,%
			height=1.4183in,%
			at={(0.758in,0.481in)},%
			scale only axis,%
			xmin=0,%
			xmax=35,%
			ymax=0.5,%
			ymin=-0.5,%
			xmajorgrids,%
			ymajorgrids,%
			ylabel style={yshift=0cm}, 
			xlabel={Time [\si{\second}]},%
			ylabel={Velocity [\si{\meter\per\second}]},%
			axis background/.style={fill=white},%
			legend style={at={(0.425,0.175)},anchor=north,legend cell 
				align=left,draw=none,legend columns=-1,align=left,draw=white!15!black}
			]
			\addplot [color=blue, dotted, line width=1.15pt] 
			file{matlabPlots/vd_downsampled.txt};%
			\addplot [color=red, dashed, line width=1.15pt] 
			file{matlabPlots/vy_downsampled.txt};%
			\addplot [color=green, solid, line width=1.15pt] 
			file{matlabPlots/vx_downsampled.txt};%
			\legend{$v_D$, $v_y$, $v_x$};%
			\end{axis}
			\end{tikzpicture}
		}
	\end{center}
	\caption{Velocity plot.}
	\label{fig:plot_velocity}
\end{figure}
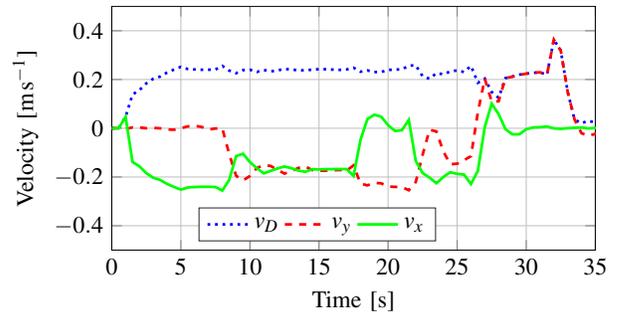

Figure~\ref{fig:plot_velocity} depicts the drone velocity $v_x$ and $v_y$ along the $x$- and 
$y$-axis, respectively, and the norm of the drone velocity $v_D$. As described in 
Sec.~\ref{sec:pathPlanner} and detailed in Appendix~\ref{appendix}, the norm of the drone velocity 
remains approximately constant while following the path. The presence of spikes might be due to the 
coupling effects of the drone $xy$ dynamics even though $x_w$ and $y_w$ references have not been 
modified yet (see, Figure~\ref{fig:plot_trajectory}). Such coupling effects are probably caused by 
the asymmetric positioning of the rotors with respect to the principal axis and the effect of the 
discrete image pixelization. 



\section{Conclusion}
\label{sec:conclusions}

In this paper, a prize-winner algorithm designed for a path following problem within the IFAC2020 
Mathworks Minidrone Competition has been presented. In particular, a lightweight and easy of 
implementation solution was set to generate the spatial coordinates of the~\ac{VTP} point to 
fulfill the competition requirements successfully. Numerical simulations carried out in MATLAB 
together with the MathWorks Virtual Reality toolbox and the Parrot support package for Simulink 
demonstrated the validity and the effectiveness of the proposed approach. The software has been 
released as open-source making it possible to go through any part of the system and to replicate 
the obtained results. Future work will include the integration of more challenging features, such 
as obstacle avoidance and 3-D reference generation, and lead to field experiments. Furthermore, the 
effects of the drone dynamics on the performance of the vision-based system will be also explored. 



%
%
%


\bibliographystyle{IEEEtran}
\bibliography{bib.bib}



\begin{appendices}
	
	\section{}
	\label{appendix}
	
	Let us consider a continuous-time dynamical system $\pazocal{H}$ and its discrete time version 
	$x_{k+1}=f(x_k,u_k)$, where $x_k, x_{k+1} \in X \subset \mathbb{R}^n$ are the current state and 
	the next state of the system, respectively, $u \in U \subset \mathbb{R}^m$ is the control 
	input. Let us consider the~\ac{PP} algorithm implementation detailed in 
	Algorithm~\ref{alg:pathPlanner}. Hence, the next state of the system $x_{k+1}$ and $y_{k+1}$ 
	along the $x$- and $y$-axis can be written as follows:
	\begin{equation}
		x_{k+1} = x_k + \alpha e_{x_k}, \qquad y_{k+1} = y_k + \alpha e_{y_k},
	\end{equation}
	respectively. After some simple algebra, we can write:
	\begin{equation}
		\dfrac{x_{k+1}-x_k}{T_{\rm PP}} = \dfrac{\alpha e_{x_k}}{T_{\rm PP}}, \qquad 
		\dfrac{y_{k+1}-y_k}{T_{\rm PP}} = \dfrac{\alpha e_{y_k}}{T_{\rm PP}}, 
	\end{equation}
	and hence,
	\begin{equation}
		v_x \approx \dfrac{\alpha e_{x_k}}{T_{\rm PP}} = \tilde{\alpha} e_{x_k}, \qquad 
		v_y  \approx  \dfrac{\alpha e_{y_k}}{T_{\rm PP}} = \tilde{\alpha} e_{y_k},
	\end{equation}
	with $\tilde{\alpha} = \nicefrac{\alpha}{T_{\rm PP}}$.
	
	Knowing that $e_{x_k}$ and $e_{y_k}$ are by definition the projections over a circle along the 
	$x$- and $y$-axis of the~\ac{VTP} with an angle $\vartheta_k$, we can write
	\begin{equation}
		\begin{array}{rll}
			e_{x_k} &=& \dfrac{R_\mathrm{max}+R_\mathrm{min}}{2} \sin{\vartheta_k},\\[10pt]
			e_{y_k} &=& \dfrac{R_\mathrm{max}+R_\mathrm{min}}{2} \cos{\vartheta_k},
		\end{array}
	\end{equation}
	and thus,
	\begin{equation} 
		\begin{split}
			V_D & = \sqrt{v_x^2 + v_y^2} \approx \tilde{\alpha} 
			\dfrac{R_\mathrm{max}+R_\mathrm{min}}{2}. \\
		\end{split}
	\end{equation}
	Hence the parameter $\alpha$ controls the drone velocity. 
	
	\hspace*{\fill} $\blacksquare$ 
	
\end{appendices}

\end{document}